\documentclass[table]{ai2style/ai2}

\usepackage{microtype}
\usepackage{hyperref}
\usepackage{url}
\usepackage{booktabs}
\usepackage{graphicx}
\usepackage{lineno}
\usepackage{enumitem}
\usepackage{listings} %
\usepackage{svg}

\definecolor{ darkblue}{rgb}{0, 0, 0.5}
\hypersetup{colorlinks=true, citecolor=darkblue, linkcolor=darkblue, urlcolor=darkblue}

\usepackage{amssymb}
\usepackage{multirow}
\usepackage{bigdelim}
\usepackage{todonotes}
\usepackage{longtable}
\usepackage{tabularray}
\usepackage{wrapfig}
\usepackage[most]{tcolorbox} 
\usepackage{url}
\usepackage{xspace}
\usepackage{svg}
\usepackage[absolute]{textpos} %

\usepackage{fdsymbol}   %

\usepackage[utf8]{inputenc} %
\usepackage[T1]{fontenc}    %
\usepackage{url}            %
\usepackage{booktabs}       %
\usepackage{amsfonts}       %
\usepackage{nicefrac}       %
\usepackage{microtype}      %
\usepackage[table]{xcolor}         %
\usepackage{amsmath}
\usepackage[most]{tcolorbox}
\usepackage{csquotes}

\usepackage{siunitx}
\usepackage{graphicx}
\usepackage{arydshln}
\usepackage{wrapfig}
\usepackage{enumitem}
\usepackage{soul} %

\usepackage{multirow}
\usepackage{xspace}
\usepackage{adjustbox}
\usepackage{pifont}
\usepackage{caption}
\usepackage{makecell}
\usepackage{subcaption}
\usepackage{bold-extra}
\usepackage{url}

\usepackage{pgf-pie}

\usepackage{hyperref}
\definecolor{linkcolor}{RGB}{0, 0, 128}
\hypersetup{
     colorlinks   = true,
     citecolor    = linkcolor,
     linkcolor    = linkcolor,
     urlcolor     = linkcolor,
}
\usepackage{pifont}%
\usepackage{listings}

\setlist[itemize]{leftmargin=*,itemsep=0em,parsep=0.3em,topsep=0.3em}

\DeclareUnicodeCharacter{2212}{\ensuremath{-}}

\definecolor{maroon}{HTML}{F26035}
\definecolor{yellow}{HTML}{FDBC42}
\definecolor{lavender}{HTML}{734f96}
\definecolor{darkergrey}{HTML}{444444}
\definecolor{midgrey}{HTML}{e6eded}
\definecolor{ai2pink}{HTML}{f0529c}%
\definecolor{ai2midpink}{HTML}{fad3e5}
\definecolor{ai2lightpink}{HTML}{fbecf3}
\definecolor{ai2midwhite}{HTML}{f2e5d9}
\definecolor{ai2offwhite}{HTML}{fbf4ee}
\definecolor{ai2green}{HTML}{0fcb8c}
\definecolor{ai2lightgreen}{HTML}{e7f9f3}
\definecolor{ai2darkgreen}{HTML}{105257}
\definecolor{ai2purple}{HTML}{B932EB}
\definecolor{ai2lightpurple}{HTML}{f7e8fc}
\definecolor{neutralEight}{HTML}{343434}
\definecolor{neutralFive}{HTML}{838383}
\definecolor{neutralThree}{HTML}{bebebe}
\definecolor{neutralOne}{HTML}{dedede}
\definecolor{lightgrey}{HTML}{fafcfc}

\usepackage{tikz}

\definecolor{maroon}{HTML}{F26035}
\definecolor{yellow}{HTML}{FDBC42}
\definecolor{darkred}{RGB}{156, 39, 33}
\definecolor{darkblue}{RGB}{31, 90, 153}
\definecolor{forestgreen}{rgb}{0.13, 0.55, 0.13}
\definecolor{olmoDarkBlue}{HTML}{012e59}
\definecolor{olmoBlue}{HTML}{265ed4}
\definecolor{olmoLightBlue}{HTML}{012e59}
\definecolor{olmoTeal}{HTML}{00d5ff}
\definecolor{olmoYellow}{HTML}{ffbb00}
\definecolor{olmoOrange}{HTML}{ff9100}

\usepackage{setspace}

\usepackage{nicematrix}
\newcolumntype{L}[1]{>{\raggedright\let\newline\\\arraybackslash\hspace{0pt}}m{#1}}
\newcolumntype{C}[1]{>{\centering\let\newline\\\arraybackslash\hspace{0pt}}m{#1}}
\newcolumntype{R}[1]{>{\raggedleft\let\newline\\\arraybackslash\hspace{0pt}}m{#1}}
\newcolumntype{P}[1]{>{\centering\let\newline\\\arraybackslash\columncolor{ai2lightpink}}m{#1}}
\newcommand{\aitwo}{\raisebox{-1.5pt}{\includegraphics[height=1.05em]{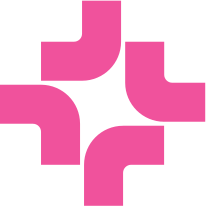}}\xspace}

\newcommand{\olmOCRLogoWithText}{\raisebox{-.7em}{\rlap{\raisebox{.8em}{\hspace{1.1em}\Large olmOCR 2}}\includegraphics[height=2.9em]{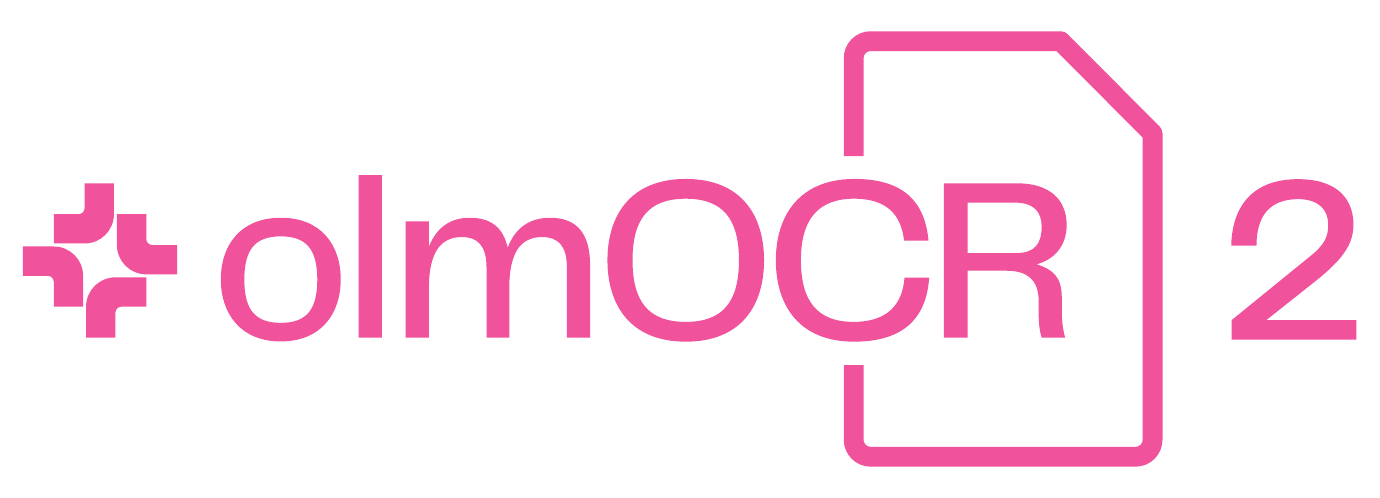}}\hspace{.1em}\xspace}

\title{\olmOCRLogoWithText\vspace{.3em}\\Unit Test Rewards for Document OCR}

\newcommand{\olmocr}{\textsc{olmOCR}\xspace}
\newcommand{\olmocrtoo}{\textsc{olmOCR 2}\xspace}

\newcommand{\bench}{\textsc{olmOCR-Bench}\xspace}
\newcommand{\model}{\texttt{olmOCR-2-7B-1025}\xspace}
\newcommand{\train}{\texttt{olmOCR-mix-0225}\xspace}
\newcommand{\traintoo}{\texttt{olmOCR-mix-1025}\xspace}

\newcommand{\synthtrain}{\texttt{olmOCR2-synthmix-1025}\xspace}

\newcommand{\huggingface}{\raisebox{-1.5pt}{\includegraphics[height=1.05em]{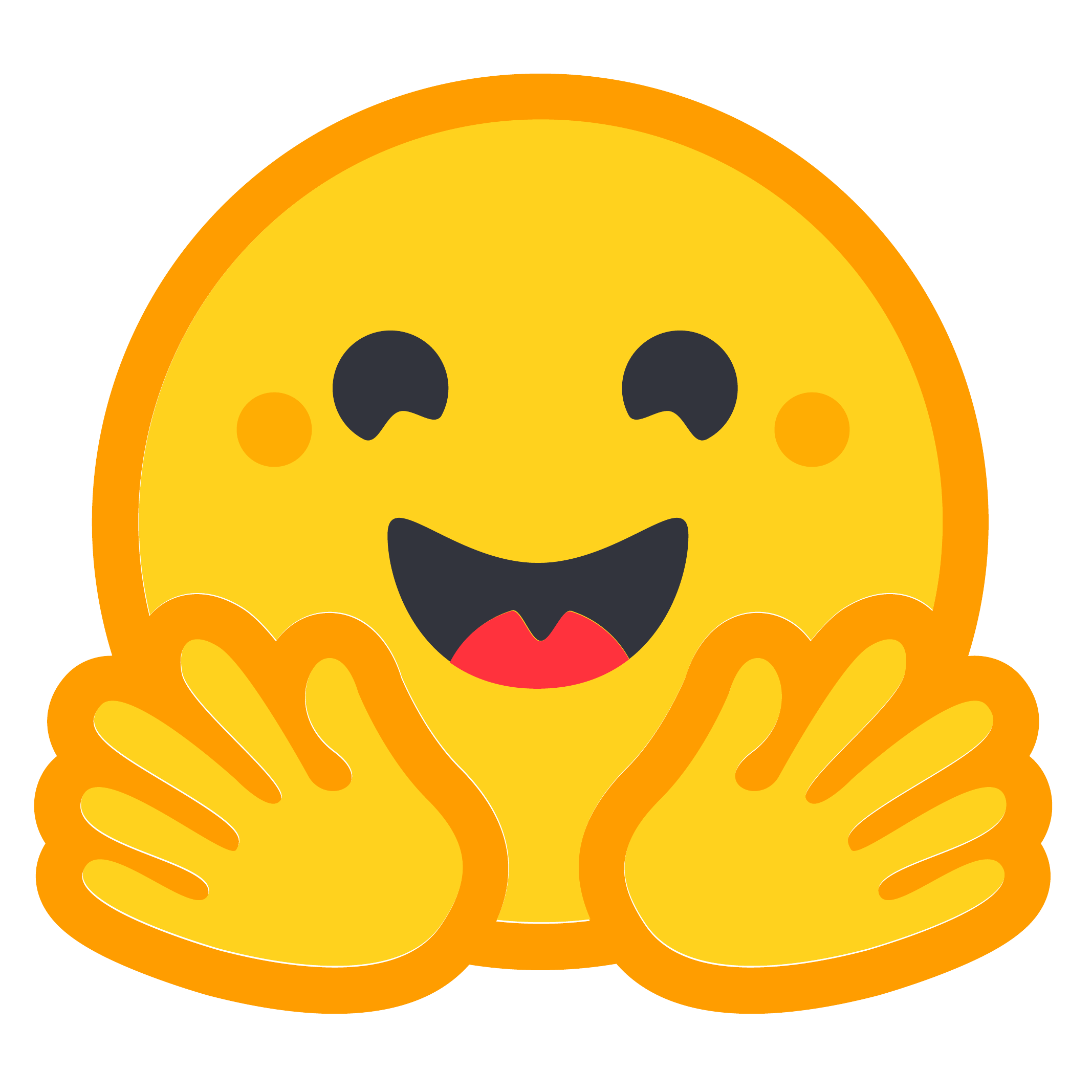}}\xspace}

\newcommand{\github}{\raisebox{-1.5pt}{\includegraphics[height=1.05em]{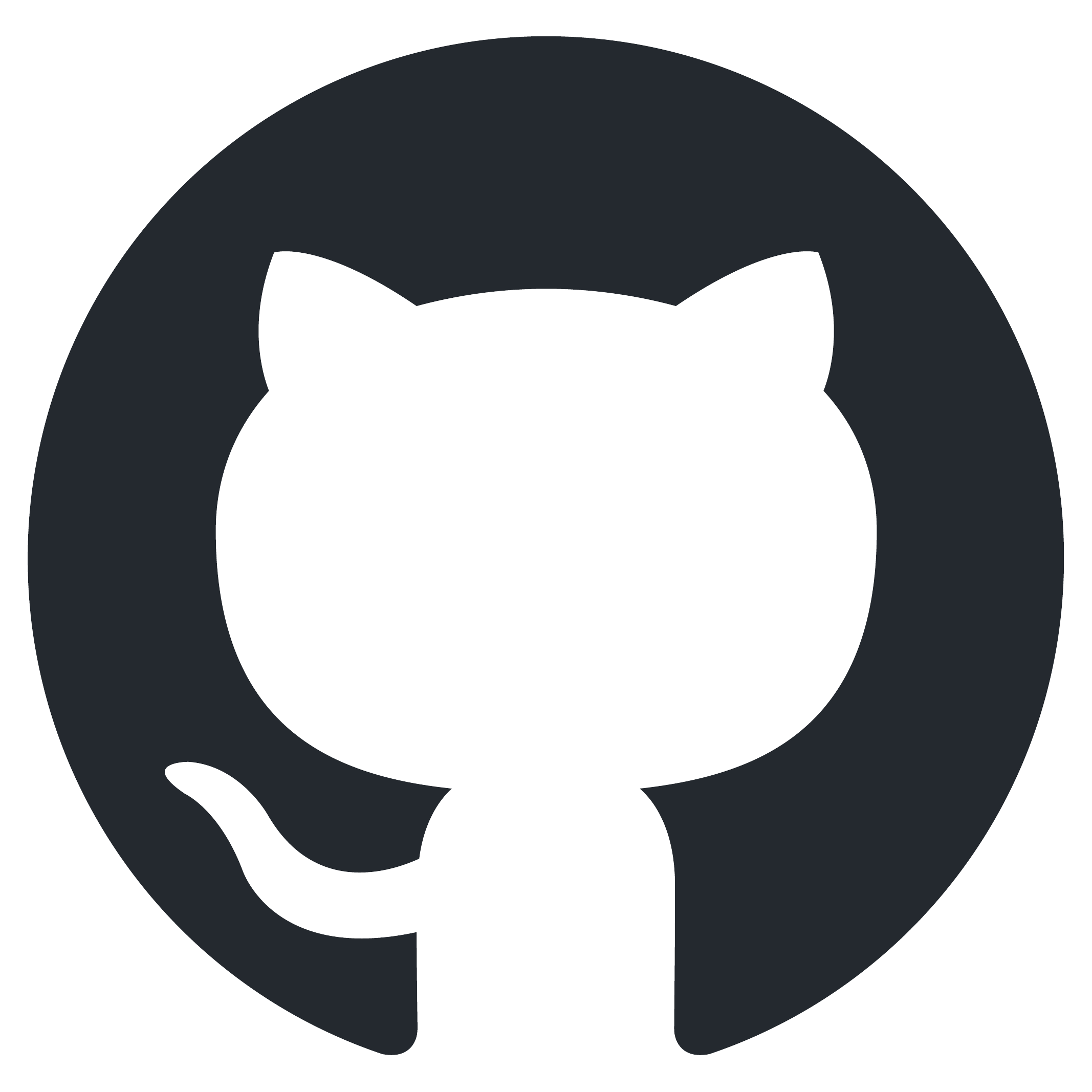}}\xspace}

\newcommand{\circleCheck}{\raisebox{-1.5pt}{\includegraphics[height=1.05em]{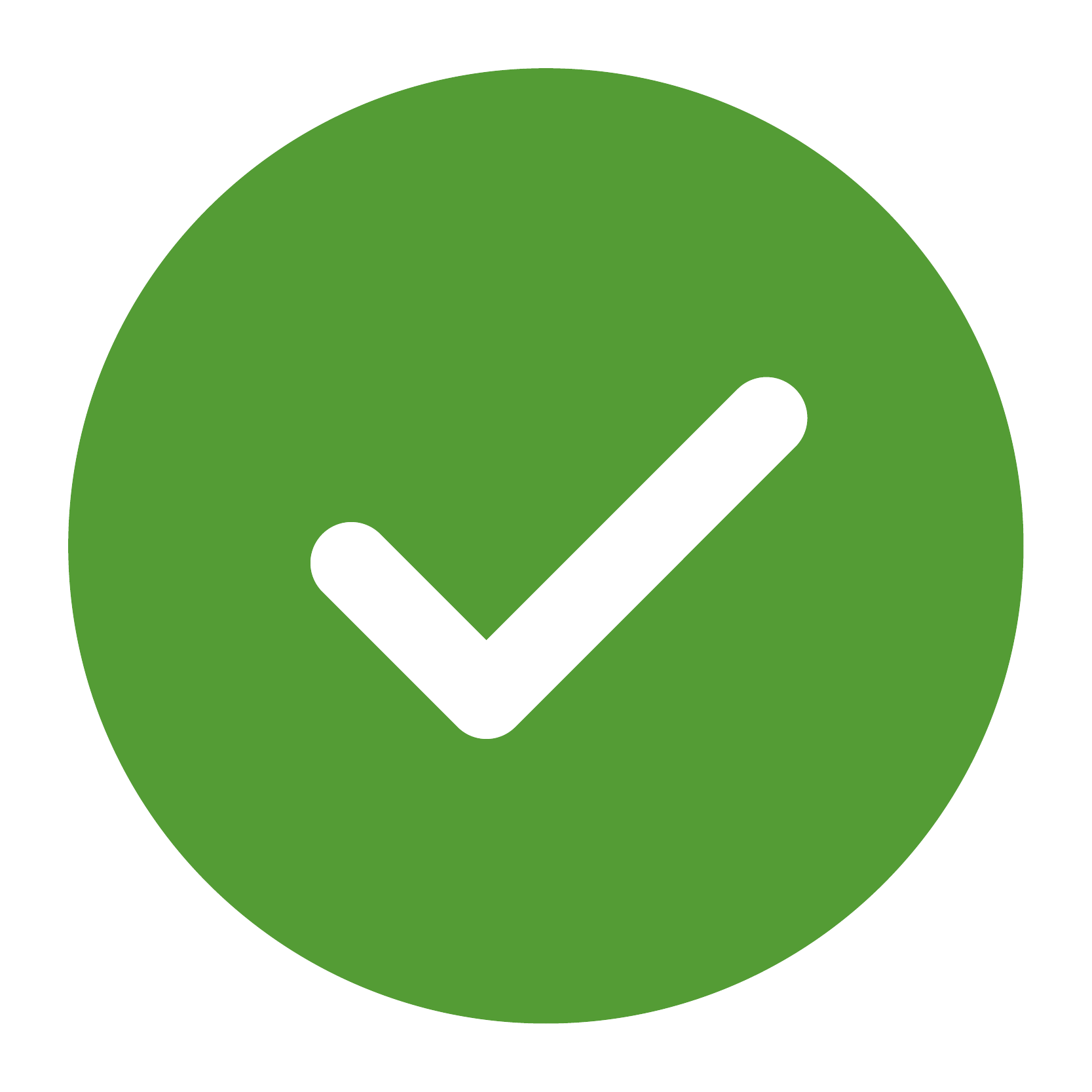}}\xspace}
\newcommand{\circleError}{\raisebox{-1.5pt}{\includegraphics[height=1.05em]{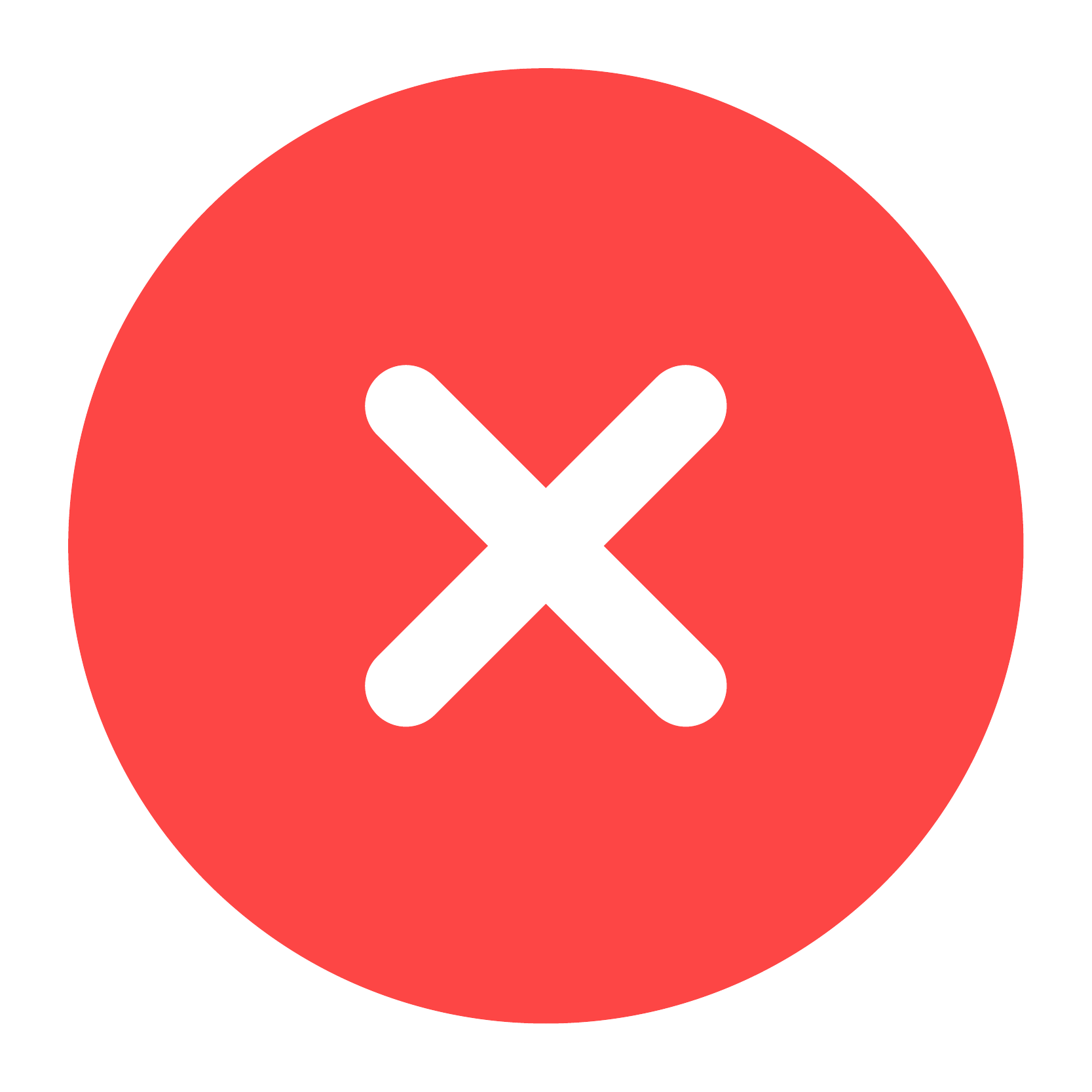}}\xspace}
\newcommand{\circleWarn}{\raisebox{-1.5pt}{\includegraphics[height=1.05em]{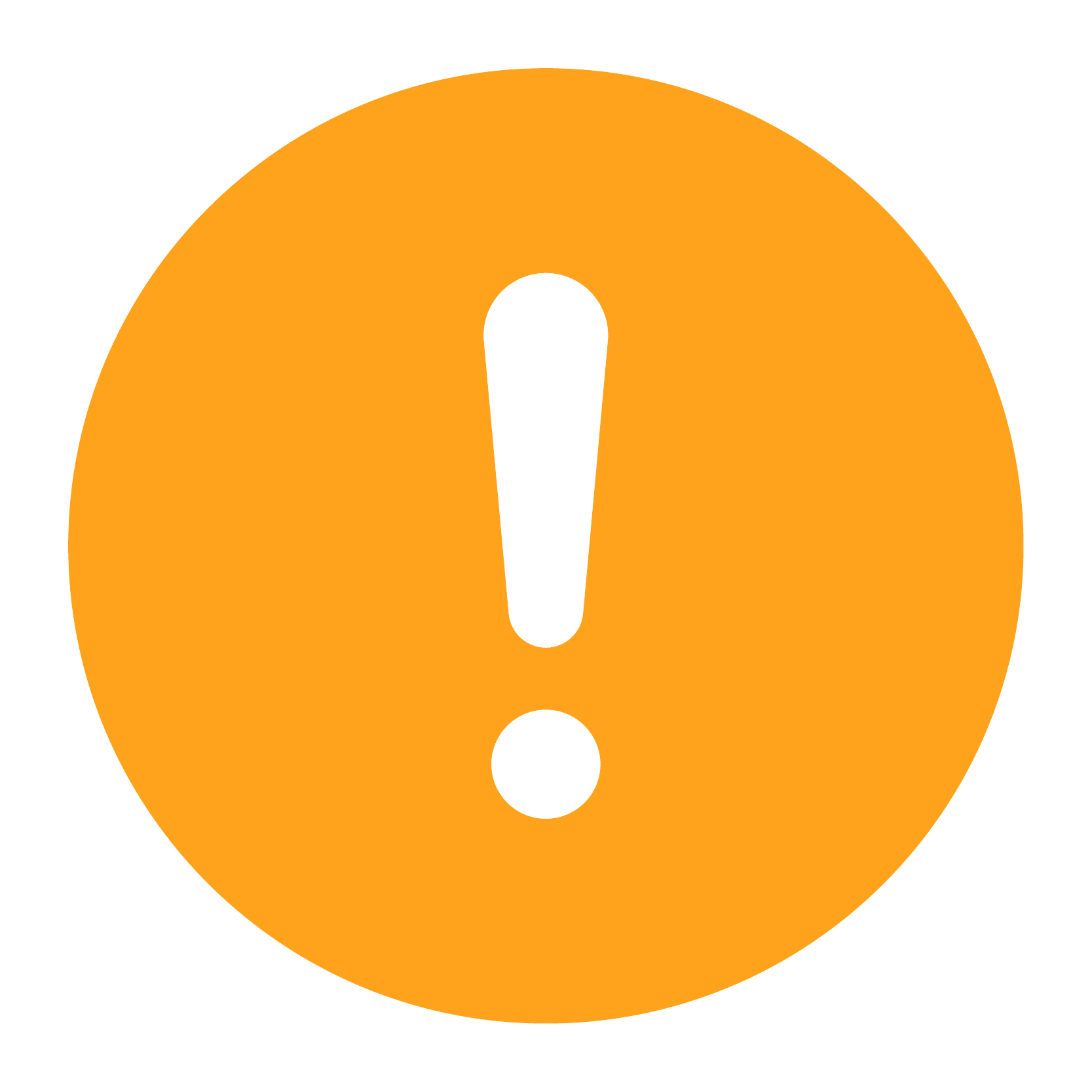}}\xspace}
\newcommand{\circleQuestion}{\raisebox{-1.5pt}{\includegraphics[height=1.05em]{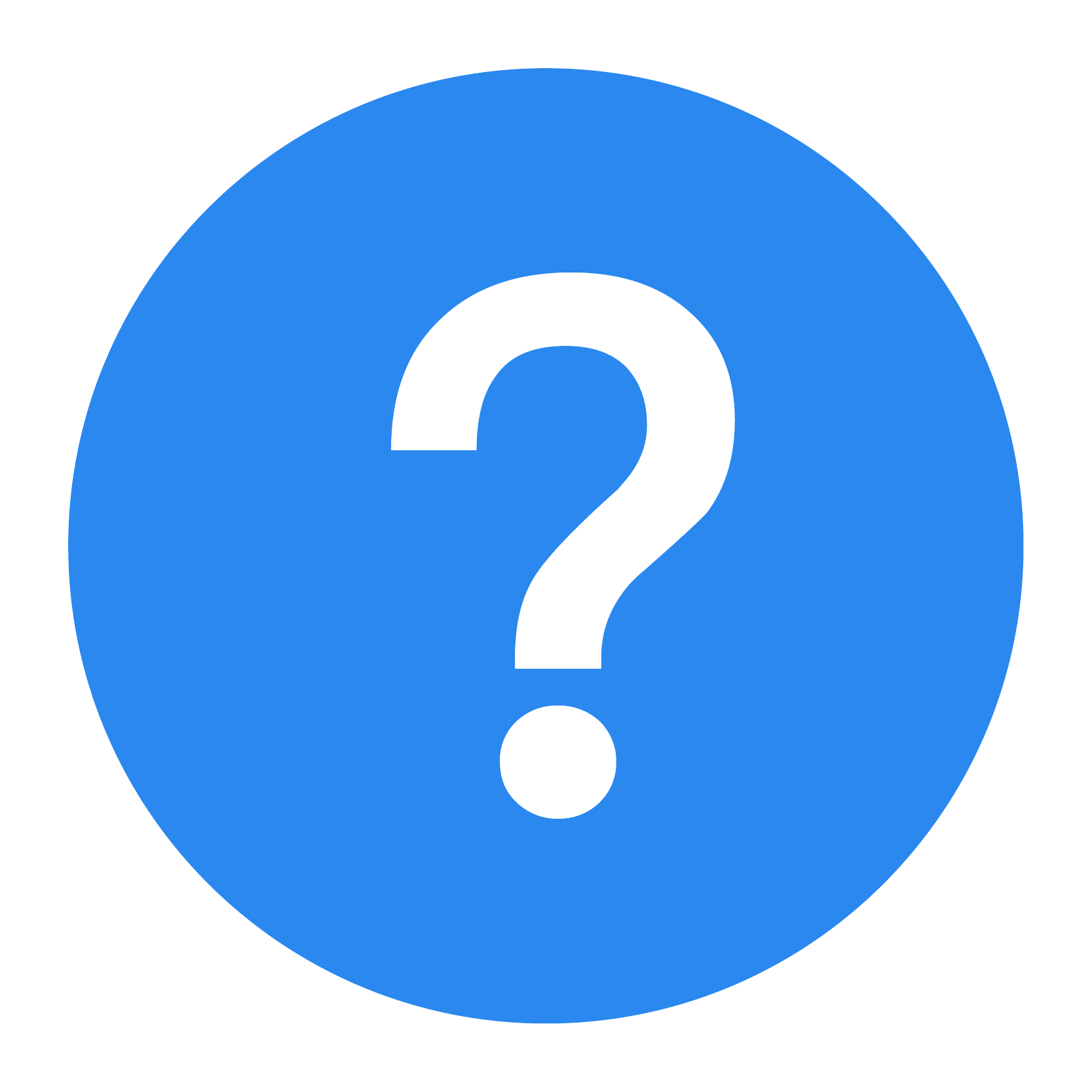}}\xspace}

\authorOne[]{Jake Poznanski}
\authorOne[]{Luca Soldaini}
\authorOne[]{Kyle Lo}

\contribution[]{Allen Institute for AI \qquad \texttt{\{jakep|lucas|kylel\}@allenai.org}}

\abstract{
We present \olmocrtoo, the latest in our family of powerful OCR systems for converting digitized print documents, like PDFs, into clean, naturally ordered plain text.
\olmocrtoo is powered by \model, a specialized, 7B vision language model (VLM) trained using reinforcement learning with verifiable rewards (RLVR), where our rewards are a diverse set of binary unit tests.
To scale unit test creation, we develop a pipeline for generating synthetic documents with diverse and challenging layouts, known ground-truth HTML source code, and extracted test cases.
We show that RL training on these test cases results in state-of-the-art performance on \bench, our English-language OCR benchmark, with the largest improvements in math formula conversion, table parsing, and multi-column layouts compared to previous versions.
We release our model, data and code under permissive open licenses.
}

\setmaintable{
\begin{table}[!h]
    \centering
    \begin{tabular}{c}
        \multicolumn{1}{c}{
            \github~{\sans{Code}}~
            \href{https://github.com/allenai/olmocr}{\texttt{allenai/olmocr}}
        } \\[4pt]
        \multicolumn{1}{c}{
            \aitwo~{\sans{Demo}}~
            \href{https://olmocr.allenai.org/}{\texttt{olmocr.allenai.org}}
        } \\[4pt]
        \multicolumn{1}{c}{
            \huggingface~{\sans{Data}}~
            \href{https://huggingface.co/datasets/allenai/olmOCR-mix-1025}{\texttt{olmOCR-mix-1025}} \quad
            \href{https://huggingface.co/datasets/allenai/olmOCR-synthmix-1025}{\texttt{olmOCR-synthmix-1025}}
        } \\[4pt]
        \multicolumn{1}{c}{
            \huggingface~{\sans{Models}}~
            \href{https://huggingface.co/allenai/olmOCR-2-7B-1025}{\texttt{olmOCR-2-7B-1025}} \quad
            \href{https://huggingface.co/allenai/olmOCR-2-7B-1025-FP8}{\texttt{olmOCR-2-7B-1025-FP8}}
        }
    \end{tabular}
\end{table}
}

\begin{document}

\maketitle

\section{Introduction}

Since our initial release of \olmocr~\citep{olmocr} in February 2025, we've seen an explosion of progress in advancing the state-of-the-art in optical character recognition (OCR).
In this short technical report, we present our latest system---\textbf{\olmocrtoo}---a state-of-the-art OCR system for extracting and linearizing content from digitized print documents like PDFs.
\olmocrtoo is powered by \model, an OCR-specialized VLM trained using reinforcement learning with verifiable rewards (RLVR)~\citep{lambert2024tulu3}.
Our training recipe involves two parts:

\begin{enumerate}
    \item We develop a synthetic document pipeline that can take any standard document, render a version of it into clean HTML, and generate easily verifiable {\bf unit tests} which can be run to check whether an OCR system output has correctly parsed this document page.
    \item We apply Group Relative Policy Optimization (GRPO)~\citep{shao2024deepseekmathpushinglimitsmathematical} to \texttt{olmOCR} using our synthetic verifiable unit tests as binary-valued reward signals.
\end{enumerate}

\noindent Others have also demonstrated the power of RLVR for OCR-specialized VLMs~\citep{wang2025infinityparserlayoutaware}; we find this training process is highly effective when combined with binary unit tests, with particular efficiency in improving the model's ability to extract equations, tables and multi-column layouts.
Combined with other performance improvements that we've made to the underlying inference system---improved base model, tuned inference settings, model checkpoint averaging or ``souping''~\citep{matena2022mergingmodelsfisherweightedaveraging,modelsoups}, bugfixes and more---\olmocrtoo achieves state-of-the-art performance on \bench{}, with a \textbf{+14.2 point overall improvement} over our initial release six months prior (Table~\ref{tab:overview}). 
Our development process over these six months has remained \textbf{fully open}, with frequent version updates accompanied by full data, model and code releases, all under permissive open source licenses.

\begin{table}[!t]
    \centering
    \begin{tabular}{lccccccc}
        \toprule
        \addlinespace[0.2ex]
        & \textbf{\makecell{\bench{}\\score}} 
        & \textbf{\makecell{Release\\date}} 
        & \textbf{\makecell{Model\\weights}} 
        & \textbf{\makecell{Training\\data}} 
        & \textbf{\makecell{Training\\code}} 
        & \textbf{\makecell{Inference\\code}} 
        & \textbf{\makecell{Model\\license}} 
        \\
        \addlinespace[0.2ex]
        \midrule
        \addlinespace[0.2ex]
        OpenAI GPT-4o 
        & $\phantom{*}68.9\pm1.1\phantom{*}$ 
        & May 2024 
        & $\phantom{^{\dagger}}\text{\circleError}\phantom{^{\dagger}}$
        & \circleError
        & \circleError
        & $\phantom{^{\dagger}}\text{\circleError}\phantom{^{\dagger}}$
        & $\text{\circleError}^\text{6}$ 
        \\ 
        Qwen 2 VL 7B 
        & $\phantom{*}31.5\pm0.9\phantom{*}$
        & Aug 2024
        & $\phantom{^{\dagger}}\text{\circleCheck}\phantom{^{\dagger}}$
        & \circleError
        & \circleError
        & $\phantom{^{\dagger}}\text{\circleCheck}\phantom{^{\dagger}}$
        & $\text{\circleCheck}^\text{1}$ 
        \\
        Gemini Flash 2 
        & $\phantom{*}57.8\pm1.1\phantom{*}$ 
        & Dec 2024 
        & $\phantom{^{\dagger}}\text{\circleError}\phantom{^{\dagger}}$
        & \circleError
        & \circleError
        & $\phantom{^{\dagger}}\text{\circleError}\phantom{^{\dagger}}$
        & $\text{\circleError}^\text{6}$ 
        \\ 
        Qwen 2.5 VL 7B 
        & $\phantom{*}65.5\pm1.2\phantom{*}$
        & Feb 2025
        & $\phantom{^{\dagger}}\text{\circleCheck}\phantom{^{\dagger}}$
        & \circleError
        & \circleError
        & $\phantom{^{\dagger}}\text{\circleCheck}\phantom{^{\dagger}}$
        & $\text{\circleCheck}^\text{1}$ 
        \\
        Mistral OCR API 
        & $\phantom{*}72.0\pm1.1\phantom{*}$
        & Mar 2025
        & $\phantom{^{\dagger}}\text{\circleError}\phantom{^{\dagger}}$
        & \circleError
        & \circleError
        & $\phantom{^{\dagger}}\text{\circleError}\phantom{^{\dagger}}$
        & $\text{\circleError}^\text{6}$ 
        \\ 
        MinerU~\texttt{1.3.10} 
        & $\phantom{*}61.5\pm1.1\phantom{*}$ 
        & Apr 2025 
        & $\phantom{^{\dagger}}\text{\circleCheck}\phantom{^{\dagger}}$
        & \circleError
        & $\phantom{^{\dagger}}\text{\circleError}\phantom{^{\dagger}}$
        & \circleCheck
        & $\text{\circleWarn}^\text{4}$
        \\
        Nanonets OCR S
        & $\phantom{*}64.5\pm1.1\phantom{*}$
        & Jun 2025 
        & $\phantom{^{\dagger}}\text{\circleCheck}\phantom{^{\dagger}}$
        & \circleError
        & \circleError
        & $\phantom{^{\dagger}}\text{\circleCheck}\phantom{^{\dagger}}$
        & $\text{\circleQuestion}^\text{5}$ 
        \\
        MonkeyOCR Pro 3B 
        & $\phantom{*}75.8\pm1.0\text{*}$ 
        & Jun 2025 
        & $\phantom{^{\dagger}}\text{\circleCheck}\phantom{^{\dagger}}$
        & \circleError
        & \circleError
        & $\phantom{^{\dagger}}\text{\circleCheck}\phantom{^{\dagger}}$
        & $\text{\circleQuestion}^\text{5}$ 
        \\
        Infinity-Parser 7B
        & $\phantom{*}79.1\pm\text{?}\text{*}\phantom{\text{??}}$ 
        & Jun 2025 
        & $\phantom{^{\dagger}}\text{\circleCheck}\phantom{^{\dagger}}$
        & \circleCheck
        & \circleError
        & $\phantom{^{\dagger}}\text{\circleCheck}\phantom{^{\dagger}}$
        & $\text{\circleCheck}^\text{1}$ 
        \\
        dots.OCR 
        & $\phantom{*}79.1\pm1.0\text{*}$ 
        & Jul 2025 
        & $\phantom{^{\dagger}}\text{\circleCheck}\phantom{^{\dagger}}$
        & \circleError
        & \circleCheck
        & $\phantom{^{\dagger}}\text{\circleCheck}\phantom{^{\dagger}}$
        & $\text{\circleCheck}^\text{2}$ 
        \\
        Marker~\texttt{1.10.1} 
        & $\phantom{*}76.1\pm1.1\phantom{*}$
        & Sep 2025 
        & $\phantom{^{\dagger}}\text{\circleCheck}\phantom{^{\dagger}}$
        & \circleError
        & $\phantom{^{\dagger}}\text{\circleError}\phantom{^{\dagger}}$
        & \circleCheck
        & $\text{\circleWarn}^\text{3}$
        \\
        MinerU~\texttt{2.5.4} 
        & $\phantom{*}75.2\pm1.1\text{*}$ 
        & Sep 2025 
        & $\phantom{^{\dagger}}\text{\circleCheck}\phantom{^{\dagger}}$
        & \circleError
        & \circleError
        & $\phantom{^{\dagger}}\text{\circleCheck}\phantom{^{\dagger}}$
        & $\text{\circleWarn}^\text{4}$
        \\
        PaddleOCR-VL 
        & $\phantom{*}80.0 \pm 1.0\text{*}$
        & Oct 2025
        & $\phantom{^{\dagger}}\text{\circleCheck}\phantom{^{\dagger}}$
        & \circleError
        & \circleCheck
        & $\phantom{^{\dagger}}\text{\circleCheck}\phantom{^{\dagger}}$
        & $\text{\circleCheck}^\text{1}$ 
        \\
        Nanonets OCR2 3B
        & $\phantom{*}69.5\pm1.1\phantom{*}$
        & Oct 2025 
        & $\phantom{^{\dagger}}\text{\circleCheck}\phantom{^{\dagger}}$
        & \circleError
        & \circleError
        & $\phantom{^{\dagger}}\text{\circleCheck}\phantom{^{\dagger}}$
        & $\text{\circleQuestion}^\text{5}$ 
        \\
        DeepSeek-OCR 
        & $\phantom{*}75.7\pm1.0\phantom{*}$
        & Oct 2025
        & $\phantom{^{\dagger}}\text{\circleCheck}\phantom{^{\dagger}}$
        & \circleError
        & \circleError
        & $\phantom{^{\dagger}}\text{\circleCheck}\phantom{^{\dagger}}$
        & $\text{\circleCheck}^\text{2}$ 
        \\
        Infinity-Parser 7B 
        & $\phantom{*}\mathbf{82.5\pm\text{\bf?}}\text{*}\phantom{\text{\bf??}}$ 
        & Oct 2025 
        & $\phantom{^{\dagger}}\text{\circleCheck}\phantom{^{\dagger}}$
        & \circleError
        & \circleError
        & $\phantom{^{\dagger}}\text{\circleCheck}\phantom{^{\dagger}}$
        & $\text{\circleCheck}^\text{1}$ 
        \\
        Chandra OCR~\texttt{0.1.0} 
        & $\phantom{*}\mathbf{83.1\pm0.9}\text{*}$
        & Oct 2025 
        & \circleCheck
        & \circleError
        & \circleError
        & \circleCheck
        & $\text{\circleWarn}^\text{3}$
        \\
        \addlinespace[0.2ex]
        \midrule
        \addlinespace[0.2ex]
        \olmocr{}
        & $\phantom{*}68.2\pm1.1\phantom{*}$
        & Feb 2025 
        & $\phantom{^{\dagger}}\text{\circleCheck}\phantom{^{\dagger}}$
        & \circleCheck
        & \circleCheck
        & $\phantom{^{\dagger}}\text{\circleCheck}\phantom{^{\dagger}}$
        & $\text{\circleCheck}^\text{1}$ 
        \\
        \olmocrtoo{} 
        & $\phantom{*}\mathbf{82.4\pm1.1}\phantom{*}$
        & Oct 2025 
        & $\phantom{^{\dagger}}\text{\circleCheck}\phantom{^{\dagger}}$
        & \circleCheck
        & \circleCheck
        & $\phantom{^{\dagger}}\text{\circleCheck}\phantom{^{\dagger}}$
        & $\text{\circleCheck}^\text{1}$ 
        \\
        \addlinespace[0.2ex]
        \bottomrule
    \end{tabular}
    \caption{
    Comparison of \olmocrtoo{} and other OCR systems. \olmocrtoo{} achieves state-of-the-art performance while maintaining fully open data, model, and code.
    Open-source licenses: $^\text{1}\text{Apache 2.0}$, $^\text{2}\text{MIT}$; 
    open licenses with usage restrictions: $^\text{3}\text{OpenRAIL-M}$, $^\text{4}\text{AGPL v3}$;
    $^\text{5}\text{license not specified}$;
    $^\text{6}\text{API access only after accepting ToS}$.
    Results are fully reproduced by ourselves, except those marked with * which are reported by their authors.
    }
    \label{tab:overview}
\end{table}

\section{Why Unit Tests?}
\label{sec:motivation}

In \bench{}~\citep{olmocr}, we measured the performance of OCR systems by defining a set of unit test cases for each document. 
These test cases can check for any of the following properties:

\begin{itemize}
    \item \textbf{Text Presence:} Checks that certain phrases appear exactly in the document
    \item \textbf{Text Absence:} Checks that certain phrases do not appear (e.g., headers, footers, or page numbers)
    \item \textbf{Natural Reading Order:} Checks sentences for reading order correctness
    \item \textbf{Table Accuracy:} Checks the relative position of cells (with specific values) in a table
    \item \textbf{Math Formula Accuracy:} Checks that a given math formula visually renders the same way with KaTeX
    \item \textbf{Baseline Robustness:} Checks that long repeated $n$-grams or non-target language characters do not appear.
\end{itemize}

\begin{figure}[!t]
    \centering
    \includegraphics[width=\linewidth,trim=1cm 1cm 1cm 1cm]{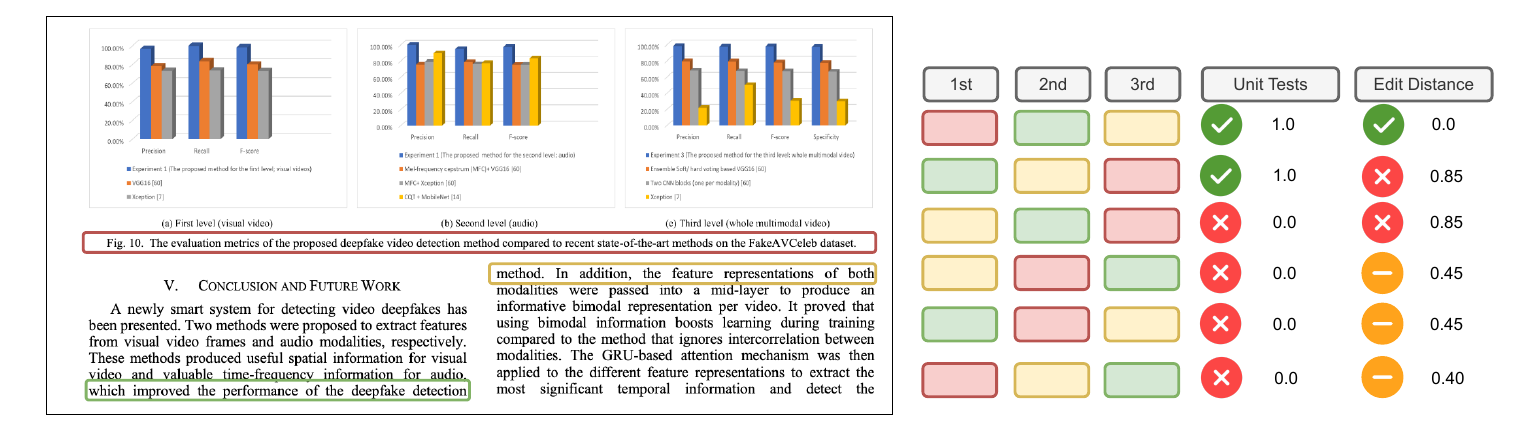}
    \vspace{0.2em} %
    \caption{Binary unit test vs edit distance for reading order errors. The \colorbox[HTML]{F1D0CD}{caption} is floating and can be correctly represented either before or after the section that contains the \colorbox[HTML]{D9E7D6}{green} and \colorbox[HTML]{FDF3D0}{yellow} passages. 
    A unit test that checks the presence of text ordering \texttt{``green, then yellow, uninterrupted by red''} will place an equivalent score to OCR output that places caption before or after the main passage. 
    Yet, edit distance highly penalizes cases where the caption occurs \emph{after} the yellow text.
    Furthermore, edit distance sometimes partially rewards cases which should be considered a severe reading order failure, such as when the caption occurs \emph{in-between} the green and yellow texts or the green then yellow text ordering is flipped.}
    \label{fig:edit-distance-reading}
\end{figure}

\begin{figure}
    \centering
    \includegraphics[width=1.0\linewidth]{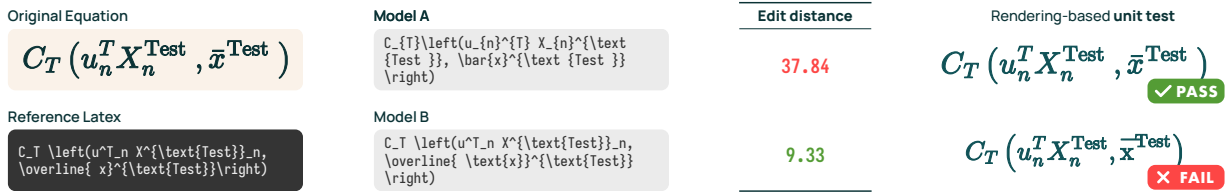}
    \caption{Binary unit test vs edit distance for math equation parsing. For a given equation and its reference LaTeX, model A produces a text output that is more dissimilar to the reference LaTeX than model B; however, after rendering and comparing the relative bounding box positions of rendered equation DOM elements, model A passes the unit test, while model B fails. Limitations of edit distance for math formulas are explored further in CDM~\citep{cdm}.}
    \label{fig:math-unit-test}
\end{figure}

While popular OCR benchmarks often use a form of edit distance~\citep{omnidocbench} against a ground truth, we developed \bench{} around \textbf{binary unit tests} for two key properties:
\begin{itemize}
    \item \textbf{Equal treatment of ``ties''.} Floating document elements like tables or figures lack a definitive ground truth representation.
    Unit tests can allow for these different-yet-equivalently-correct representations of the same OCR'd content to yield similar scores, while edit distance often rewards/penalizes these cases differently.
    \item \textbf{Continuous score doesn't necessarily measure ``correctness''.} The use of edit distance as a continuous scoring function rewards/penalizes OCR output in a manner that doesn't correlate with practical notions of correctness, such as placing greater emphasis on correct ordering of main body text rather than caption placement or post-rendered correctness of a LaTeX formula rather than the LaTeX form itself.
\end{itemize}
\indent We include two key motivating examples in Figures~\ref{fig:edit-distance-reading} and \ref{fig:math-unit-test} to further illustrate.

While prior work has explored improvements to edit distance, particularly for math formulas~\citep{cdm}, and such ideas have led to recent updates in popular benchmarks like OmniDocBench v1.5~\citep{omnidocbench}, 
there is still much more work to be done to develop calibrated continuous scores for other types of OCR targets beyond math formulas. 
Binary unit tests, on the other hand, offer us a single elegant framework to simultaneously develop evaluations for a diversity of OCR errors.

\section{Scaling Unit Test Generation for RLVR}

\begin{figure}[t]
    \centering
    \includegraphics[width=1.0\linewidth]{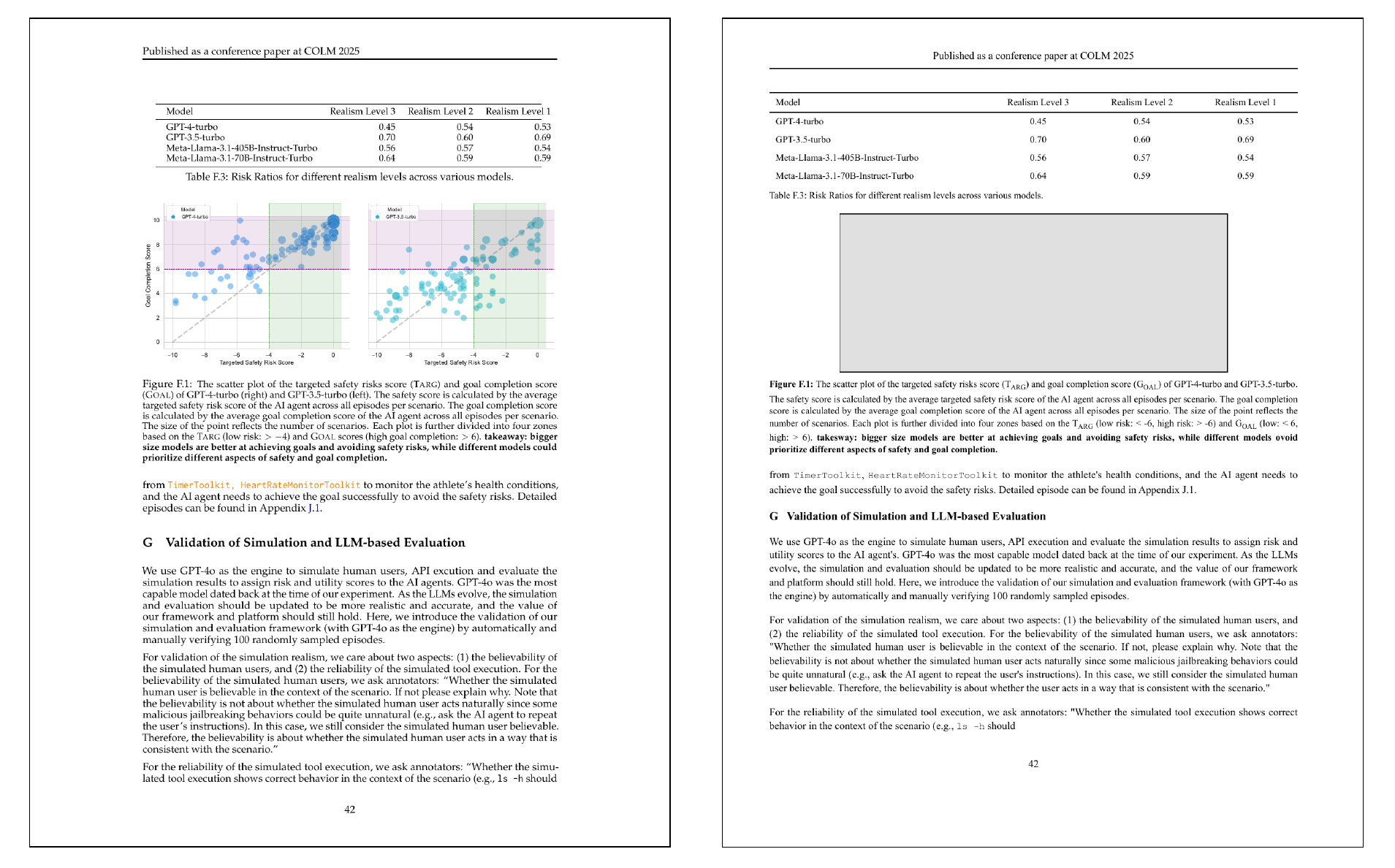}
    \caption{HTML page generation for our \olmocrtoo synthetic data pipeline. We sample a page from a real document (left) and prompt a general VLM to generate a highly similar HTML page (right). The rendered HTML page image paired with the raw HTML serves as supervision for our OCR-specialized VLM.}
    \label{fig:html-render}
\end{figure}

\subsection{Data}
The original unit tests that make up \bench{} were all manually verified and took hours of work
to create and check by hand. 
In order to scale unit test creation to support RL training, we develop a pipeline to create large numbers of synthetic test cases with very high accuracy.
The pipeline synthetically creates HTML pages corresponding to real PDF documents, which allows programmatic generation of unit tests. 
An example of the generated HTML is shown in Figure~\ref{fig:html-render}.

\paragraph{PDF sourcing} We sample documents that contain relevant, difficult-to-OCR material. For example, to focus on unit tests for math equations, we source from arXiv math-heavy papers. By sampling real-world documents, we create a high diversity of documents, instead of being restricted to just a handful of pre-made templates. 

\paragraph{PDF to HTML conversion} We iteratively prompt a general VLM to first create, and then refine, the HTML code that best represents the rasterized image of a page. 
In detail, this can be broken down in three steps:

\begin{enumerate}
\item\textbf{Layout analysis}. We first use the VLM with a picture of a randomly sampled page from PDF documents and ask it to analyze the document. In this step, we prompt\footnote{\href{https://github.com/allenai/olmocr/blob/f5fad405c0bc47ce7196fad5b9f2c69d33da4ef2/olmocr/bench/synth/mine_html_templates.py\#L398-L420}{\path{github.com/allenai/olmocr/olmocr/bench/synth/mine_html_templates.py#L398-L420}}} the VLM to identify the general layout of the page, such as number of columns, presence of images or tables, headers and footers, and so on. This step provides guidance during HTML page generation to improve coverage of unit test elements. 
\item\textbf{Content rendering}. We prompt\footnote{\href{https://github.com/allenai/olmocr/blob/f5fad405c0bc47ce7196fad5b9f2c69d33da4ef2/olmocr/bench/synth/mine_html_templates.py\#L437-L465}{\path{github.com/allenai/olmocr/olmocr/bench/synth/mine_html_templates.py#L437-L465}}} the general VLM again with the previous model output and the same document image, and ask it to ``render this document as clean, semantic HTML'' fitting into the same dimensions as the original.
\item\textbf{Output refinement}. We render the HTML generated at the previous step, convert it to an image, and pass it to the general VLM along with the original document image and the generated HTML. We prompt\footnote{\href{https://github.com/allenai/olmocr/blob/f5fad405c0bc47ce7196fad5b9f2c69d33da4ef2/olmocr/bench/synth/mine_html_templates.py\#L510-L546}{\path{github.com/allenai/olmocr/olmocr/bench/synth/mine_html_templates.py#L510-L546}}} the general VLM to refine its HTML to better match the original.
\end{enumerate}

\paragraph{Unit test creation} We create \bench{}-compatible test cases based on the semantics of the HTML the VLM produced. For example, the layout analysis step asks for headers and footers to be in HTML \texttt{<header>} and \texttt{<footer>} tags, so we can generate ``Text Absence'' test cases for those. Math equations are rendered with KaTeX, so we can extract those and create test cases matching them. Tables are extracted from the ground-truth in the same way, and random cells sampled to create test cases.

\paragraph{Implementation} We use \texttt{claude-sonnet-4-20250514} as the general VLM for the procedure described above. 
Overall, we found it sufficiently accurate and cost effective, costing approximately \$0.12 per document page.
We note that our pipeline is robust to hallucinations:
even in cases where Claude makes an error when it is performing OCR, that does not affect our pipeline, as we use the HTML output alone to generate unit tests.
\synthtrain, our final data mix consists of 2,186 PDF pages.
In total, across these PDF pages, we create 30,381 test cases.

Alongside \synthtrain, we use a refreshed mix for supervised fine-tuning, \traintoo.
The dataset contains 267,962 pages from over 100,000 PDFs sampled from diverse sources, including 9,828 pages from national archives. 
Compared to \train, the new mix has been re-processed using GPT-4.1 instead of GPT-4o, has more consistent equation formatting (with \texttt{\textbackslash[} and \texttt{\textbackslash(} for block and inline math), uses HTML format for tables, and includes basic alt text for images.
See Table~\ref{tab:sft-data-change} for SFT results using these two training sets.

\begin{table}[!h]
\centering
\small
\setlength{\tabcolsep}{3pt} %
\begin{tabular}{p{9em}*{8}{c}@{\hspace{2em}}c} %
\toprule
\addlinespace[0.4ex]
&
\textbf{\makecell{ArXiv}} &
\textbf{\makecell{Old\\scans\\math}} &
\textbf{\makecell{Tables}} &
\textbf{\makecell{Old\\scans}} &
\textbf{\makecell{Headers\\\& footers}} &
\textbf{\makecell{Multi\\column}} &
\textbf{\makecell{Long\\tiny\\text}} &
\textbf{\makecell{Base}} &
\textbf{\makecell{Overall}} \\
\addlinespace[0.2ex]
\midrule
\train & 78.6 & 79.9 & 72.9 & 43.9 & 95.1 & 77.3 & 81.2 & 98.9 & 78.5 $\pm$ 1.1 \\
\traintoo & 70.8 & 79.3 & 77.9 & 45.6 & 93.7 & 81.3 & 78.7 & 99.3 & 78.3 $\pm$ 1.2 \\
\addlinespace[0.2ex]
\bottomrule
\end{tabular}
\caption{Finetuning on a single epoch of \train vs \traintoo, evaluated on \bench.}
\label{tab:sft-data-change}
\end{table}

\subsection{Training}

\begin{figure}[t]
    \centering
    \includegraphics[width=1.0\linewidth]{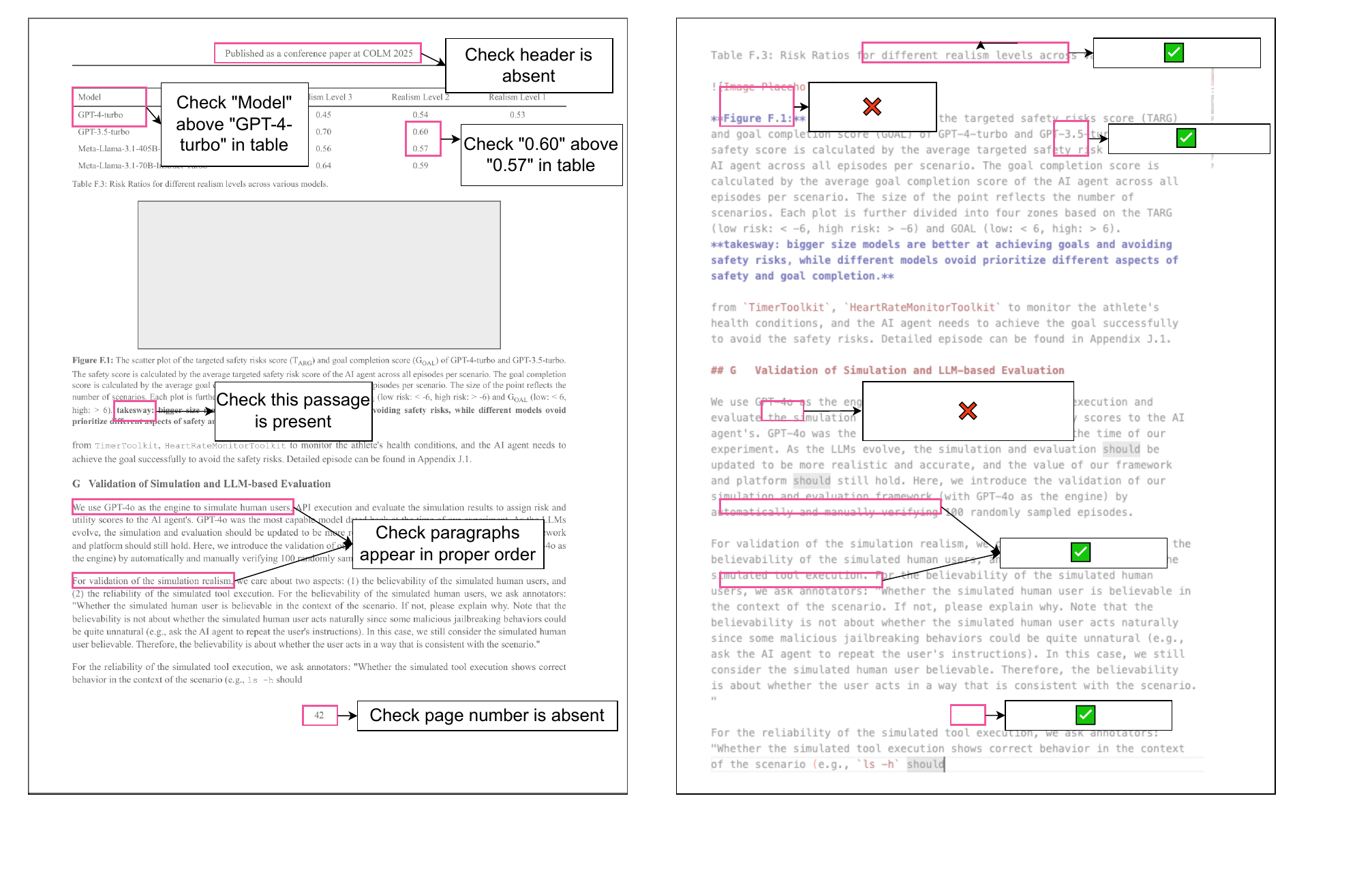}
    \caption{Unit test rewards for \olmocrtoo's RLVR training. Given a generated HTMl page and its unit tests (left), we can easily score a generated Markdown page (right) according to these unit tests. Each test contributes a binary reward which is aggregated at a page-level as a pass rate. For example, with 4 of 6 passes, the page level reward is 0.67.}
    \label{fig:grpo-reward}
\end{figure}

We start with a Qwen2.5-VL-7B-Instruct model that has been fine-tuned on \traintoo as described in \citet{olmocr}.
We train for one epoch on \synthtrain using an 8xH100 GPU node. For each document, 28 completions are generated. Each completion gets scored using the standard \bench{} scoring rules, where each test case is either a pass or fail, and the reward is the fraction from 0.0 to 1.0 of passing test cases. An example of this reward is shown in Figure~\ref{fig:grpo-reward}.

Besides the unit test above, we include two additional rewards to ensure correct output format: 
a binary reward for whether the model completion ends with the EOS token, and a reward between 0 and 1 to ensure that the model outputs document metadata at the top of its response (\textit{e.g.}, primary language, rotation correction factor).

We use the Hugging Face TRL library~\citep{vonwerra2022trl}, with KL divergence $\beta = 0.01$. 
To maximize performance, we found it beneficial to train multiple models, and average, or \textit{soup}~\citep{modelsoups}, their weights.
In detail, we train six models with different random seeds, and soup their weights at the end.

\begin{table}[t]
\centering
\small
\setlength{\tabcolsep}{3pt} %
\begin{tabular}{p{14em}*{8}{c}@{\hspace{2em}}{c}} %
\toprule
\addlinespace[0.4ex]        
&
\textbf{\makecell{ArXiv}} & 
\textbf{\makecell{Old\\scans\\math}} & 
\textbf{\makecell{Tables}} & 
\textbf{\makecell{Old\\scans}} & 
\textbf{\makecell{Headers\\\& footers}} & 
\textbf{\makecell{Multi\\column}} & 
\textbf{\makecell{Long\\tiny\\text}} & 
\textbf{\makecell{Base}} & 
\textbf{\makecell{Overall}} \\
\addlinespace[0.2ex]
\midrule
\addlinespace[0.2ex]
Mistral OCR API & 77.2 & 67.5 & 60.6 & 29.3 & 93.6 & 71.3 & 77.1 & 99.4 & $ 72.0\pm1.1 $ \\
Marker~\texttt{1.10.1} & 83.8 & 66.8 & 72.9 & 33.5 & 86.6 & 80.0 & 85.7 & 99.3 & $ 76.1\pm1.1 $ \\
MinerU~\texttt{2.5.4}* & 76.6 & 54.6 & 84.9 & 33.7 & 96.6 & 78.2 & 83.5 & 93.7 & $ 75.2\pm1.1 $ \\
DeepSeek-OCR & 77.2 & 73.6 & 80.2 & 33.3 & 96.1 & 66.4 & 79.4 & 99.8 & $ 75.7\pm1.0 $ \\
Nanonets-OCR2-3B & 75.4 & 46.1 & 86.8 & 40.9 & 32.1 & 81.9 & 93.0 & 99.6 & $ 69.5\pm1.1 $ \\
PaddleOCR-VL* & 85.7 & 71.0 & 84.1 & 37.8 &  97.0 & 79.9 & 85.7 & 98.5 & $ 80.0\pm1.0 $ \\
Infinity-Parser 7B* & 84.4 & 83.8 & 85.0 &  47.9 & 88.7 & 84.2 & 86.4 & 99.8 & $ 82.5\pm\text{?}\phantom{\text{??}}$ \\	
Chandra OCR \texttt{0.1.0}* & 82.2 & 80.3 & { 88.0} & {50.4} & 90.8 & 81.2 & 92.3 & {99.9} & $83.1\pm0.9$ \\	
\addlinespace[0.2ex]
\midrule
\addlinespace[0.2ex]
\olmocr{} (first release) & 63.3 & 67.5 & 62.3 & 38.6 & 93.4 & 67.6 & 54.8 & 97.9 & $ 68.2\pm1.1 $ \\
\addlinespace[0.2ex]
$+$ Dynamic temp scaling & 71.4 & 73.1 & 65.6 & 40.5 & 93.2 & 76.6 & 64.9 & 96.7 & $ 72.8\pm1.2 $ \\
\addlinespace[0.2ex]
$+$ Better prompting & 76.3 & 76.0 & 70.2 & 43.2 & 94.1 & 77.5 & 71.9 & 96.8 & $ 75.8\pm1.0 $ \\
\addlinespace[0.2ex]
$+$ \makecell[l]{New trainer, YAML,\vspace{-.25em}\\ img resize, Qwen 2.5 VL} & 78.8 & 77.5 & 71.9 & 45.4 & 94.2 & 78.6 & 81.4 & 99.8 & $ 78.5\pm1.1 $ \\
\addlinespace[0.2ex]
$+$ Handle blank pages & 78.6 & 79.9 & 72.9 & 43.9 & 95.1 & 77.3 & 81.2 & 98.9 & $ 78.5\pm1.1 $ \\
\addlinespace[0.2ex]
$+$ \makecell[l]{Synth data, \vspace{-.25em}\\ RLVR, souping}  & 83.0 & 82.3 & 84.9 & 47.7 & 96.1 & 83.7 & 81.9 & 99.7 & $82.4\pm1.1$ \\
\addlinespace[0.2ex]        
\bottomrule
\end{tabular}
\caption{OCR model performance comparison. Results are reproduced in-house, except those marked with *, which are reported by model authors.}
\label{tab:ocr_comparison}
\end{table}

\section{Results}

Table~\ref{tab:ocr_comparison} presents a summary of major development points between our initial \olmocr{} and \olmocrtoo{}, evaluated on the latest version of \bench{}.
We also include a number of powerful OCR baselines, including the latest versions of actively developed open OCR projects like Marker, MinerU, and PaddleOCR, as well as some recent additions to the state-of-the-art in OCR-specialized VLMs.
Our key findings are:

\paragraph{Dynamic temperature scaling.} Our first version of \olmocr{} set a default temperature of 0.8. We found that sampling at a lower temperature tends to give better results but at the risk of VLM inference encountering repetition loops. 
To take advantage of low temperatures while mitigating this repetition issue, we use dynamic temperature scaling starting at 0.1 and continually increasing it to 0.2, 0.3 and so on up to a max of 0.8. 
Each increase is triggered off a failure in the model to generate an EOS token (and thus repeat infinitely).
This resulted in significant improvement in overall benchmark performance.

\paragraph{Better prompting.} We found an unintended bug in which order of image and the text was mismatched between training and inference prompts. 
We standardize prompt order by always including text first in all settings; matching the order in training and inference improved benchmark performance substantially. 
We experimented with the reverse order and found no meaningful difference in OCR performance, however placing any fixed text first allows for prompt caching by the inference engine.

\paragraph{New trainer.} We reimplemented our trainer for VLM finetuning, with minor tweaks to hyperparameters (\textit{e.g.}, avoiding weight decay on the bias and layer norm weights). We found no meaningful benchmark score difference from this change.

\paragraph{YAML.} The first \olmocr{} was trained to output JSON objects. We switched to YAML, which reduced the retry rate dramatically. We speculate this is because the model does not need to remember how many open quotes there are currently in the JSON and can simply output an EOS token as soon as it is done. 
With JSON, we also found more incidences of repetition loops. 
We found no benchmark score difference, but with fewer need for retries, this improved our inference efficiency.

\paragraph{Image Resizing.} Our initial \olmocr{} used 1024px on the longest edge; \olmocrtoo{} uses 1288px instead. 
Bigger images do appear to yield slightly better performance across many model families, though they take more dedicated compute. 
We performed a sweep of image sizes and picked this size as a reasonable balance between benchmark score and inference speed.

\paragraph{Qwen 2.5 VL.} We switched from Qwen 2 VL~\citep{wang2024qwen2vlenhancingvisionlanguagemodels}, which was our base model in \olmocr{} to Qwen 2.5 VL~\citep{qwen25vl}, resulting in a slight improvement in benchmark score.

\paragraph{Handle blank pages.} We caught a bug in the data loader for our \olmocr{} model where all instances of blank pages were being skipped.
The model, never having been trained on blank pages, would hallucinate in such cases. 
We fixed the data loader and retrained the model, though this didn't impact benchmark scores.

\paragraph{\olmocrtoo.} Finally, our latest release \olmocrtoo{} demonstrates a significant improvement in benchmark performance. Our best model, reported here, is the result of:

\begin{enumerate}
    \item A single epoch of SFT training on \traintoo,
    \item A single epoch of RL training over our synthetic data \synthtrain,
    \item Repeating the RL training for six random seeds and averaging (or ``souping'') the checkpoints.
    We used importance sampling at both the token level (3 runs) and the sequence level (3 runs); more details on their difference in \cite{zheng2025groupsequencepolicyoptimization}.
\end{enumerate}

\section{Related Work}
\label{sec:related-work}
\paragraph{Machine learning models for OCR.} OCR of digitized print documents, often in PDF format, has been a long-standing research area, even dating back to the 1950s~\citep{Mori1992-qy,Smith2013-pp}; these systems were largely built on hand-written pipelines based on expert understanding of the PDF internal representation~\citep{pdfa2015}.
The incorporation of modern machine learning models into these pipelines marked a notable paradigm shift, leading to the development of powerful OCR systems like MinerU~\citep{mineru}, Marker~\citep{marker} and PP-OCRv5~\citep{paddleocr}.
Such systems often compose multiple models together (\textit{e.g.}, section segmentation or table parsing using small, specialized models).

\paragraph{Rise of vision language models.} We are seeing yet another paradigm shift in OCR methodology, relying on increasing power of vision language models (VLMs) to generate the target OCR text in an end-to-end fashion.
This was a rare pattern prior to 2025. 
Notable exceptions to the rule include Nougat~\citep{nougat} and GOT-OCR 2.0~\citep{gottheory}, models capable of taking images of PDF pages as input and return plain text.
Also in 2024 was the release of GPT-4o~\citep{openai2024gpt4technicalreport},
which boasted another major leap in PDF understanding, and we saw other frontier model developers soon after release general VLMs with improved OCR capabilities (\emph{e.g.} Gemini 2~\citep{geminiPDF} and Qwen 2.5VL~\citep{qwen25vl}).
Our initial release of olmOCR~\citep{olmocr} demonstrated the ability to distill GPT-4o's OCR capability into a small 7B VLM.
Using VLMs as the foundation for OCR has since seen widespread adoption with ever more impressive models; notable examples include end-to-end systems like Nanonets-OCR2-3B~\citep{Nanonets-OCR2}, MinerU 2.5~\citep{niu2025mineru2}, dots.OCR~\citep{jian2025dotsocr}, Monkey OCR~\citep{monkeyOCR} and Chandra OCR~\citep{paruchuri2025chandra}, as well as hybrid systems like DeepSeek-OCR~\citep{deepseek_ocr_2025} and PaddleOCR-VL~\citep{cui2025paddleocrvlboostingmultilingualdocument} that use powerful VLMs as the backbone within ML pipelines.

\paragraph{Reinforcement learning for OCR}
Several other recent models have explored reinforcement learning for OCR.
DianJin-OCR-R1~\citep{chen2025dianjinocrr1enhancingocrcapabilities} uses RL rewards to finetune a reasoning model to improve OCR performance
by using chain of thought to dedicate more inference compute to difficult document sections.
Other works such as \citep{he2025seeingbelievingmitigatingocr} and \citep{xiong2025docr1evidencepageguidedgrpo} have demonstrated that RL rewards improve performance in visual document answering systems.

The closest work to ours is Infinity Parser~\citep{wang2025infinityparserlayoutaware} which also develops a synthetic data pipeline around HTML renderings and trains their OCR-specialized VLM using GRPO with verifiable rewards.
A slight difference in our works is our use of sampled real content to seed generation of full HTML pages while their work injected sampled real content into pre-made HTML layouts.
A more significant difference is that we use binary unit tests as our verifiable reward signal while they define their reward based on edit distance, paragraph count, and structural consistency.

\section{Conclusion}
We have presented \olmocrtoo, a state-of-the-art OCR system powered by an OCR-specialized VLM trained using reinforcement learning with verifiable rewards.
We define these rewards using binary unit tests and scale the generation of these tests through a synthetic data pipeline that samples real documents and generates similar HTML renderings as ground truth.
We also present our learnings through the course of our ongoing open development of \olmocr{}.
We release our model checkpoints, training and inference code, and two training data mixes, all under permissive open licenses to support further research in this field.

In the future, we hope to further develop the synthetic data pipeline to cover more complicated document types and unit tests. We are interested in exploring further the differences between binary unit tests versus continuous scores like edit distance as evaluation targets~\citep{omnidocbench} as well as RL rewards~\citep{wang2025infinityparserlayoutaware}.

\section*{Acknowledgements}

We thank members of the Ai2 team for their support in making this release possible, especially Kyle Wiggers and Crystal Nam for their support during the release process.
We thank our inference partners DeepInfra and Parasail for helping us set up public API access to \olmocrtoo.
We thank Haydn Jones, Charitarth Chugh, and Vik Paruchuri for their contributions to our open source repo.
We thank the Qwen team for releasing open VLM models that have accelerated this exciting line of work.
We thank the many developers of other open OCR systems for their usage of and feedback on \bench{}.
This research used resources of the Oak Ridge Leadership Computing Facility, which is a DOE Office of Science User Facility supported under Contract DE-AC05-00OR22725.

\bibliographystyle{plainnat}
\bibliography{references}

\end{document}